\documentclass[runningheads]{llncs}
\usepackage[utf8]{inputenc}
\usepackage{graphicx}
\usepackage{amsmath,amsfonts,amssymb,xcolor}
\usepackage{color}

\usepackage{booktabs}
\usepackage{arydshln}
\usepackage{bbold}
\usepackage{mathtools}  
\usepackage{xfrac}  
\usepackage{soul} 
\usepackage{dsfont}
\usepackage{todonotes}
\usepackage{graphicx}
\usepackage[pagebackref=true,breaklinks=true,colorlinks,bookmarks=false]{hyperref}
\usepackage{ulem}
\usepackage{physics}

\usepackage{comment,wrapfig}
\usepackage{multirow}

\newcommand{\np}[1]{\todo[color=blue!10]{NP: #1}}

\newcommand{\Embp}{{E_\text{patch}}}
\newcommand{\embp}{{\vb*{e}_\text{patch}}}
\newcommand{\Embi}{{E_\text{image}}}
\newcommand{\embi}{{\vb*{e}_\text{image}}}
\newcommand{\Embl}{{E_\text{label}}}
\newcommand{\embl}{{\vb*{e}_\text{label}}}

\newcommand{\zp}{{z}^\text{+}}
\newcommand{\zm}{{z}^\text{-}}

\newcommand{\hy}{{\hat y}}

\newcommand{\lossname}{\text{WN}}
\newcommand{\commentaire}[1]{}

\begin{document}
%
\title{%
A patch-based architecture for multi-label classification from single label annotations 
}

\titlerunning{ }
\authorrunning{ }

\author{%
    Warren Jouanneau\textsuperscript{1,2}, %
    Aurélie Bugeau\textsuperscript{1}, %
    Marc Palyart\textsuperscript{2}, %
    Nicolas Papadakis\textsuperscript{3},\\
    Laurent Vézard\textsuperscript{2}\\
}
\institute{%
    \vspace*{-5pt}
    \textsuperscript{1}Univ.  Bordeaux, Bordeaux INP, CNRS, LaBRI, UMR 5800,F-33400 Talence, France\\
    \textsuperscript{2}Lectra, F-33610 Cestas, France\\
    \textsuperscript{3}Univ.  Bordeaux, Bordeaux INP, CNRS, IMB, UMR 5251,F-33400 Talence, France\\
}
\maketitle
\vspace*{-10pt}
\begin{abstract}
In this paper, we propose a patch-based architecture for multi-label classification problems where only a single positive label is observed in images of the dataset. 
Our contributions are twofold. First, we introduce a light patch architecture based on the attention mechanism.
Next, leveraging on patch embedding self-similarities, we provide a novel strategy for estimating negative examples  and deal with positive and unlabeled learning problems.
Experiments demonstrate that our architecture can be trained from scratch, whereas pre-training on similar databases is required for related methods from the literature.
\end{abstract}

\vspace*{-20pt}
\section{Introduction}
\vspace*{-5pt}

Data annotation, or labeling, is at the core of supervised learning approaches. In image classification, state-of-the-art methods 
rely on an ever-increasing amount of data, which makes the annotation collection a major issue. 
Training datasets are made of image-label associations obtained from an expensive manual annotation, an automatic collection, or filtering and mapping of already existing image descriptions. Assembling a dataset proves to be difficult, especially when: experts must annotate unlabeled data; rare events have to be recognized; or a dataset is created from different sources with inconsistent label taxonomies.

The accurate characterization of most images requires multi-label classification.
The content of an image is indeed generally rich in information, as it includes multiple structured components.
Associating a single label to an image is thus a too restricted setting. It is preferable to annotate an image with several and non-exclusive labels (e.g. presence/absence of wood, metal, fabric, etc.).

For supervised learning purpose, positive and negative examples for each label are needed in the training dataset. The image-labels associations need to be exhaustive. Any missing or incorrect annotation for the label $l$ on a given image $X$  leads to a wrong 
example for this label. Such errors may have an impact on the complete labeling of the image $X$ and for the characterization of the label $l$ on the remaining images of the dataset.
Naturally, obtaining  error free multi-label annotations drastically makes the dataset creation more complex.

To alleviate the task of dataset creation,  weakly supervised learning methods only rely on partial data annotations. Such methods combine approaches ranging from fully supervised to unsupervised learning, e.g. few shot learning where only a small set of example is available for each label.

A special case of weak supervision for classification is positive and unlabeled (PU) learning~\cite{bekker2020learning}. In PU learning, only partial positive labeling is available. In the multi-label PU setting, only a subset of images of the training data set are annotated for each label. For the remaining images, we have no information, which means that we do not know if the label is present or not in the image. Several labels can potentially be identified in a single image.

In the multi-label positive and unlabeled learning context, obtaining annotations is greatly simplified. PU learning is indeed adapted to automatic data collection and dataset merging. 
Positive examples for each label can be collected independently of what they represent for other labels. More generally, as it does not require negative examples and annotation completeness, PU is well suited to handle heterogeneous labeling taxonomies coming from different datasets. 


In PU learning, one difficulty is to deal with the absence of negative examples. The situation is even worse in the multi-label context, where each image is actually both positive for certain labels and negative for other ones. 

\subsubsection{Contributions and outline.}
In this work, we propose to address the negative example issue with a patch-based approach. We argue that the patch level is better suited than the image level to achieve multi-label characterization. If an image is a positive example regarding a given label, then some of its patches can be considered positives and others negatives. For positive image examples, the patch-label association is nevertheless unknown. As discussed in the related work section~\ref{sec:rel_works}, this makes the multi-label PU problem difficult to solve.

Our main contribution, presented in section \ref{sec:methodo} is a light patch-based architecture dedicated to multi-label PU learning. By considering an image as a set of patches, we first build multi-label image representations with a patch attention mechanism. Then, assuming that a small image patch contains at most one single label, we also propose to  estimate negative examples by leveraging on patch-based image representation self-similarities.

In section \ref{sec:expes}, experimental results demonstrate that our patch-based framework is adapted to multi-label classification problems with single positive annotations, while providing an explicit spatial localization of labels.
When training models from scratch, our architecture generalizes faster and better than Resnet-5, while being significantly lighter (reduction $\times 100$ of the number of parameters).

\section{Related works}\label{sec:rel_works}

In this section, we first review multi-label classification models using set of patches. Next, we discuss existing strategies to obtain a global image representation from the information contained in a set of patches and their embeddings. Finally, we present state-of-the-art  methods for positive and unlabeled  learning.



\subsection{Patch embeddings for  multi-label image classification}
Multi-label classification of images  
can be split into multiple single label classification tasks \cite{read2009classifier}, where each classifier is in charge of predicting a specific label. Recent works nevertheless demonstrate the interest of tackling the joint multi-label problem~\cite{wei2014cnn}. 
In image data, there is indeed a correlation between localization and labels. As a consequence, detection methods not only intend to infer the image labels, but they also aim at estimating their localization through bounding boxes~\cite{ren2015faster,redmon2016you} or segmentation masks~\cite{he2017mask}.

Considering an image as a set of patches is an  appropriate model for multi-label classification. 
In practical applications, unless a hierarchical label taxonomy is considered (hairs, head, body), labels are indeed related to a subpart of the image only (e.g. wood, brick, metal...). 
With a patch-based approach, each patch specializes and contributes to the classification with respect to a single label. 

While patch-based approaches have been first introduced for texture synthesis problems~\cite{EfrLeu1999}, their interest has been demonstrated for image level tasks such as denoising~\cite{BuadesCollMorel2005}, super-resolution~\cite{Freeman2002}, segmentation, labeling~\cite{Coupe2011}, classification~\cite{varma2008statistical}, etc. In particular, the representation of an image from its extracted patches has been thoroughly studied, and we refer the reader to~\cite{liu2019bow} for a review of existing methods from the bag of word framework to recent deep models. 


The Transformer architecture~\cite{vaswani2017attention}, originally proposed in the natural language processing community, has soon been transposed to image tasks. In order to adapt Transformer architecture to images, the ViT method~\cite{dosovitskiy2020image} considers an image as a set of patches or as a sequence (if patch position is encoded) of patches. 
The ViT achieves impressive results without  convolution layers.


Many relevant extensions of the image Transformer concern the construction of a relevant set of elements to feed to the network. As an example, CrossViT~\cite{chen2021crossvit} relies on patches of different size, which brings multi-scale information. 
In other works, a convolution neural network (CNN) can be used as a stem \cite{xiao2021early} or as a feature pyramid \cite{zhang2020feature} to feed a Transformer network. 


Relying on sets of patch embeddings proved to be a powerful methodology for image classification with modern learning methods. The ConvMixer model~\cite{trockman2022patches} shows for instance that Transformers' performances are mostly due to the representation of images as a set of patches, rather than to the architecture itself.
%
When facing image classification problems, one has to come back from patch level to image level at some point. This is the subject of the next section. 

\subsection{From set of patch embeddings to image representation}

We now present methods providing a relevant image representation from the embeddings of image patches. The image representation must handle multi-label classification and deal with the presence of multiple instances of these labels in the images. To that end, the information contained in the set of embedded patches has to be aggregated. 
In the literature, the aggregation of elements in a set is mainly done using pooling operator such as average, max, min or sum of the element representations~\cite{zaheer2017deep}. The pooling of feature maps related to receptive fields of different sizes is for instance usually done with global average~\cite{lin2013network}. 

The raise in popularity of the attention mechanism has led to new pooling methods~\cite{ilse2018attention} realizing weighted sums of element representations.
Transformers~\cite{vaswani2017attention}, relying on multi-head attention, can also operate as element pooling. 
The transformer uses cross-attention with weights given by a similarity score between the element representations and "queries". In the case of patches, these "queries" can be seen as codebooks~\cite{zhao2022codedvtr}.
Queries can be designed beforehand, or they can be parameters learned by the model.
A main limitation of the transformer is its quadratic complexity with respect to the dimension of input data.
In order to reduce the computational burden of transformers, the perceiver~\cite{jaegle2021perceiver} architecture  computes intermediate latent representations of reduced dimension before realizing cross-attentions with the queries. 
Hence, a task dependent element pooling can be considered for each query. The queries can therefore be defined as label embeddings~\cite{lanchantin2021general}, with an independent pooling for each label.


The element pooling in attention architectures naturally deals with multiple instances of a single label in an image. In computer vision, multiple instance learning consists in  both detecting the presence of a label at the image level and localizing accurately the corresponding instances~\cite{carbonneau2018multiple}. 
The attention mechanism provides a joint solution to these problems~\cite{ilse2018attention}.  
The detection of a label is indeed given by the pooling of the patch embeddings with the corresponding query. The similarity weight between a patch and a query indicates the degree of participation of the patch in the label decision. If a patch has an important weight in the pooling of a label predicted as positive, then this patch should contain relevant information relative to this label. As a consequence, the similarity weights can help to localize distinct subareas of the image corresponding to a single label.



In a supervised setting, the patch-based attention mechanism is thus adapted to both multi-label and multiple instance cases.
However, its application to weakly supervised problems with positive only annotations at the image level requires the development of new methods.


\subsection{Positive and Unlabeled learning for  multi-label classification}
Positive and unlabeled (PU) classification is a weakly supervised classification problem where only positive examples are available. As studied in the  review~\cite{bekker2020learning}, many methods have addressed this problem in the single-label setting. 


However, only few methods dedicated to PU learning for multi-label classification problems exist. As detailed in~\cite{cole2021multi}, most works consider the many (and not single) positive~\cite{kanehira2016multi} case,  the single positive or negative one~\cite{huang2018nonconvex} or the existence of negatives for each label~\cite{ishida2017learning}.

In this paper, we focus on the complex multi-label application case where only a single positive label is known for each element of the training set.
In~\cite{mac2019presence},  all but the known positive examples are considered as negatives and included as groundtruth negative examples in the loss function optimized during training. This corresponds to a uniform penalization of the  positive predictions, that can be enhanced with a dedicated spatial consistency loss~\cite{verelst2022spatial}.
Cole et al. \cite{cole2021multi} propose to enhance this kind of methods with the Regularized Online Label Estimation (ROLE) strategy. The ROLE model first improves  the loss function with a term penalizing the distance between the number of positive label predictions for an image and a hyperparameter. This hyperparameter corresponds to the mean number of positive labels per image that is nevertheless unknown in general. Then, an online label estimation strategy is considered for unobserved labels. It consists in learning  new parameters that should correspond to the groundtruth value of unknown labels, that can be either positive or negative. These estimations are realized in a separate branch of the model and are compared with the actual predictions of the model in a dedicated loss. 



The ROLE model makes an interesting proposition with the online estimation of negative examples. Nevertheless, it does not leverage on data content to tackle the multi-label aspect of the problem. The online strategy mainly stabilizes the learning through memorization of former label predictions. In practice, it reinforces the tendencies (positive or negative predictions) provided by the current model, by making the predicted weight values closer and closer to $0$ or $1$. 
We argue that the patch-based approach is a suitable strategy to estimate negative examples in the context of  multi-label PU learning.




\section{Methodology}\label{sec:methodo}




We start this section with a formal statement of the multi-label classification problem addressed in this paper.
Given a set of labels $l\in L$, the objective is to determine  $\mathds{P}(x_n|l)$, the probability of presence of the label $l$ in an image $x_n$.  We denote as ${\mathbf{y}}_n=\{y_{n,l}\}_{l\in L}$ the ground truth labels that indicate  if a class $l\in L$ is present ($y_{n,l}=1$) or not ($y_{n,l}=0$) in an image $x_n$.
Hence, the multi-label problem can be formulated as the estimation of a labeling score $\hat{\mathbf{y}}_n=\{\hat y_{n,l}\}_{l\in L}$, where $\hat y_{n,l}\in[0,1]$ indicates the presence ($\hat y_{n,l}\to 1$) or absence ($\hat y_{n,l}\to 0$) of the label $l$ in the image $x_n$.

Training multi-label classification in a fully-supervised manner requires entirely annotated data, for which the collection and annotation are very costly. In our case, this complete annotation is not available. We have partial ground truth annotations on an image dataset  containing partial positive ${\mathbf{z}}^+_n$ and negative ${\mathbf{z}}^-_n$ examples for the image $x_n$. We  consider that $\zp_{n,l}=1$ (resp. $\zm_{n,l}=1$) means that the label $l$ is present (resp. absent) in the annotated image $x_n$. 
If $\zp_{n,l}=\zm_{n,l}=0$ then we have no information on the label $l$. Finally, the two annotated sets are assumed compatible, so that the case $\zp_{n,l}=\zm_{n,l}=1$ is impossible.

In the Positive and Unlabeled (PU) context  considered in this paper, nothing is known about the negatives ($\zm_{n,l}=0$ for all $n$ and $l$) and the positive labels are only partially observed. 
In our applications, we nevertheless consider that one single positive label $\zp_{n,l}=1$ is available per image $x_n$.




To solve the PU multi-label classification problem, we propose a weakly supervised method able to deal with partial positive labeling. Our method relies on the use of image patches. It is built on top of a key hypothesis: a small enough image patch contains only one label.

In section~\ref{sec:archi}, we describe our patch-based architecture for multi-label classification. Section~\ref{sec:loss} presents the loss proposed to train the model. In section~\ref{sec:neg}, we build on our patch-based architecture to provide negative examples and deal with the  positive and unlabeled learning problem.

\subsection{Patch-based architecture}\label{sec:archi}
\begin{figure}[ht!]
    \includegraphics[width=1.0\linewidth]{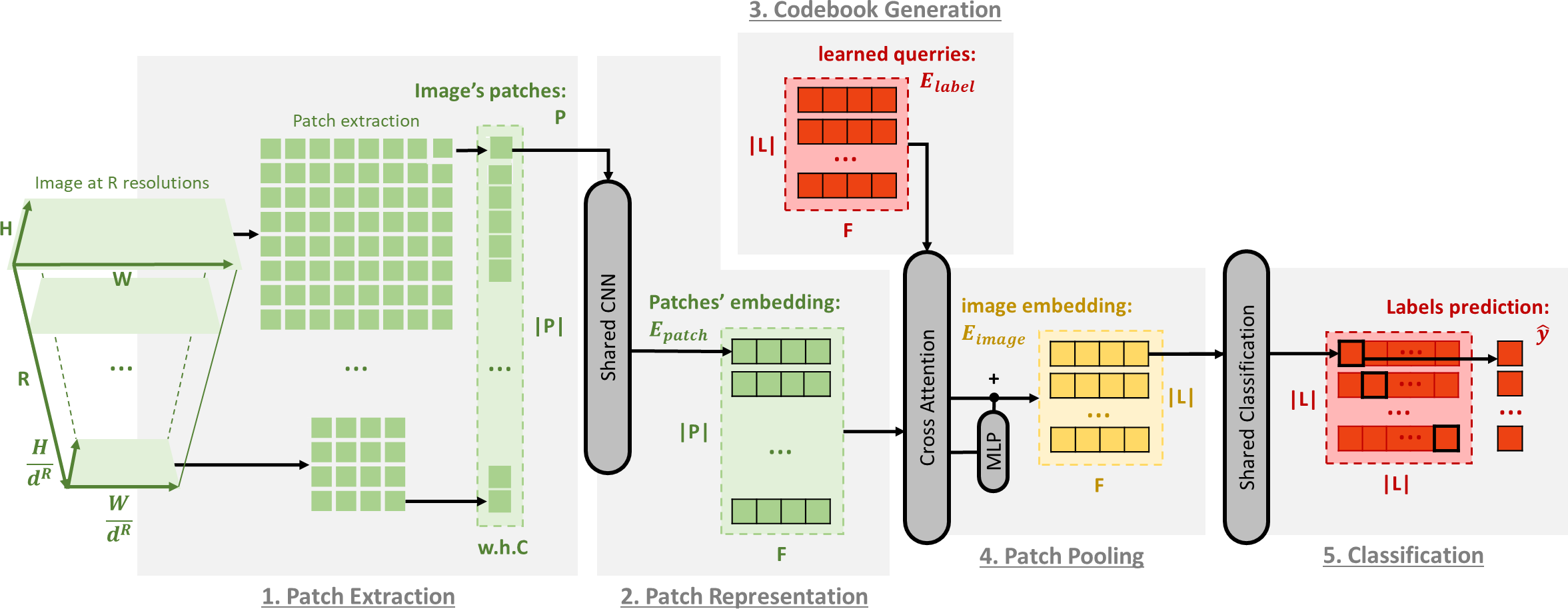}
    \caption{Proposed multi-resolution patch deep neural network architecture.}\vspace*{-0.5cm}
    \label{fig:archi}
\end{figure}

We now provide a detailed description of our patch-based architecture. 
As illustrated in Figure~\ref{fig:archi}, the architecture is composed of five blocks \ref{fig:archi} based on the bag of word framework.
A subset of patch is first selected. The patches are then embedded in a general patch representation space, in which also live label codebooks. A pooling of the patch embedding for each label is then performed using an attention mechanism. It is followed by the final multi-label classification.
%
%
%
%
\subsubsection{Patch extraction.}
(Block 1 of figure \ref{fig:archi}):
In our architecture, an image is first converted into a set of patches extracted at different resolutions. Each image $I \in \mathbb{R}^{H \times W \times C}$ of height $H$, width $W$ and with $C$ channels is downsampled spatially R times with a constant uniform ratio $d$, using bilinear interpolation.
For each resulting image, patches $P$ are extracted using a sliding window looping through the image both horizontally and vertically with a uniform window size $h \times w$ and stride. At each resolution, we denote a patch $p \in \mathbb{R}^{h \times w \times C}$.
%
The patch size being the same across all resolutions, the set of patches contain fine to coarse 
information with decreasing resolution. 
In this work, we consider the same size for the stride and the sliding windows, so that patches  form a perfect grid of the image. When the image size is too large, patches can be randomly subsampled at each resolution level to limit the computational burden.


%
%



\subsubsection{Patch representation.} 
(Block 2 of figure \ref{fig:archi}):
Each patch is fed into a CNN architecture. The weights of this model are shared across all patches. With this backbone model, all $m$ patches are projected into the same latent space of dimension $F$, $\Embp = \{\embp_i \mid \embp_i \in \mathbb{R}^{F}, i = 1\cdots m\}$,  resulting in a vector of embedded features $\embp_i \in \Embp $ that play the role of patch descriptors. The CNN architecture consists of multiple efficientNet blocks \cite{tan2019efficientnet}. 

\subsubsection{Codebook embedding.}
(Block 3 of figure \ref{fig:archi}):
Each label is associated with its own representative patch, $\embl_l$, that we call codebook. We assume that each representation $\embl_l$  contains the embedded feature that a patch should contain to be discriminated as positive in regard to the corresponding label $l$.  The set of all codebooks, denoted, $\Embl = \{\embl_l \mid \embl_l \in \mathbb{R}^{F}, l \in L\}$,  is obtained through back-propagation.

\subsubsection{Image representations with patch pooling.}
(Block 4 of figure \ref{fig:archi}):
An image representation is created from the embedded patches. To that end, we propose to consider attention pooling. The pooling is done using cross-attention between the set $\Embp$ of embedded patches and the learned label codebooks $\Embl$. 
This attention mechanism can be seen as a two steps approach. 
The first step consists in evaluating the relevance of all selected patches with respect to each possible label. To do so, we consider the scalar product between vectors to define a score matrix $A$ of weights $\alpha_{l,i}$ between the representation $\embp_i$ of patch $i$ and the representation $\embl_l$ of label $l$:
\begin{equation}\label{eq:attention-score}
    \alpha_{l,i}= \frac%
    {\exp(\embl^{}_l \: . \: \embp_i)}%
    {\sum_{j = 1 \cdots n} \exp(\embl^{}_l \: . \: \embp_j)}. 
\end{equation}


The second step then realizes a weighted sum of the patch representations through the matrix product $AE_{patch}$. Inspired from \cite{vaswani2017attention}, we finally define the image representation as:
\begin{equation}
    E_{image} = f(A E_{patch}) + A  E_{patch}, \label{eq:attention-represention}
\end{equation}
where  $f$ is a feed forward MLP. 
With this attention framework, we get multiple image representations $\Embi = \{\embi_l \mid \embi_l \in \mathbb{R}^{F}, l \in L\}$. Hence, one image representation $\embi_l$ embeds features of a subset of patches matching the global patch representation of a label $l \in  L$.



\subsubsection{Classifier.}
(Block 5 of figure \ref{fig:archi}):
We propose to realize the single multi-label classification with a shared classifier. Given an image representation $\embi_l$, a classifier provides a prediction $\hy_l$ relative to the presence of labels $l\in L$ in the image. 
In practice, the predictions are obtained with a softmax operator
\begin{equation}\label{eq:multi-classification}
    \hy_l= \frac%
    {\exp(W_l{\embi}_l)}%
    {\sum_{k \in L} \exp(W_k{\embi}_l)}, 
\end{equation}
where $W_l$ are weight matrices that are learned for each label. 
The weights being shared with the softmax operator, the classifier operates a partition of the latent space. Our objective with this classifier model is to get a label specialization of the embedding space. In other words, this classification architecture is designed to enforce each learned label codebook $\embl_l$ to be the centroid of the patch embeddings relative to the label $l$.




Overall we suggest that our architecture enforces intra cluster consistency with the patch pooling and inter cluster dissimilarity with the classifier.
Indeed, while the 
scalar product between patches representation relative to the same label is maximized by the attention pooling (see relation \eqref{eq:attention-score}), the classifier block implicitly aims at minimizing the 
scalar product between label image representations $\embi_l$.


\subsection{Training loss}\label{sec:loss}
We now present the multi-label loss used in our framework to achieve a prediction $\mathbf{\hy}$ of the ground truth $\mathbf{y}$ from positive and negative examples  $\mathbf{z}^+$ and $\mathbf{z}^-$. 
First, to introduce the different losses and avoid possible confusions, we review the differences between multi-class and multi-label problems.
We recall that we omit the image index $n$ to simplify the notations: $\mathbf{\hy}=\{\hy_l\}_{l\in L}$ is the prediction of the probability of presence of labels $l$ for any single image.

\subsubsection{Supervised multi-class learning.} In multi-class problems, classes are mutually exclusive: for each image,  there exists a single label $l$ such that $y_l=1$ in the ground truth, while $y_k=0$ for all $k\neq l$.  As a consequence, the predictions $\hy_l$ are constrained to belong to the simplex (i.e. $\hy_l\geq 0$ and $\sum_{l \in L}\hy_l=1$). 
The standard loss function is then the Cross Entropy $\mathcal{L}_{\text{CE}}$ between the available positive examples $\mathbf{z}^+$ and the normalized predictions $\hat{\mathbf{y}}$: 
\begin{equation}\label{eq:lce}
    \mathcal{L}_{\text{CE}}(\mathbf{z}^+,\hat{\mathbf{y}}) = - \sum_{l \in L} 
    z^+_l\log(\hy_l).
\end{equation}

\subsubsection{Supervised multi-label learning.}  On the other hand, in multi-label classification,  multiple positive groundtruth are possible for a single image ($\sum_{l\in L} y_l\geq 1$). Hence, the simplex constraint can not be considered anymore, and each label prediction is an independent score $\hy_l\in[0,1]$. Negative predictions must also be taken into account and compared with  negative examples contained in $\zm$.
In this work, we use the Binary Cross Entropy loss  $\mathcal{L}_{\text{BCE}}$, that is a standard loss for multi-label classification. For each class $l\in L$, BCE is the sum of two cross entropy \eqref{eq:lce} terms between positive $z_l^+=1$ (resp. negative $z_l^-=1$) ground truths observations and positive $\hat y_l$ (resp. negative $1-\hat y_l$) predictions:
\begin{equation}\label{eq:lbce}
    \mathcal{L}_{\text{BCE}}(\hat{\mathbf{y}}) = \mathcal{L}_{\text{CE}}(\mathbf{z}^+,\hat{\mathbf{y}}) + \mathcal{L}_{\text{CE}}(\mathbf{z}^-,1-\hat{\mathbf{y}}).
\end{equation}

\subsubsection{Positive and Unlabeled learning \rm{(PU)}.}    
PU is a multi-label problem that has two main difficulties: (1) the groundtruth is partially labeled and (2) we only have access to positive examples ($\zm_{l}=0$ for all $l\in L$). 

To handle the partial labeling of positive examples, we make the assumption that, using the loss $\mathcal{L}_{\text{BCE}}$, the features learned on one image for a given label will transpose globally to others.  
%
To deal with unknown negative labels, two possibilities can be distinguished.  The first one consists in using the available positive labels only and train the model with the loss $\mathcal{L}_{\text{CE}}(\mathbf{z}^+,\mathbf{\hy})$. However, this loss function is globally minimized with the trivial solution predicting all labels as positive for all images. The second  option is to consider all labels except the observed one as negative examples~\cite{mac2019presence}, i.e. training with the loss $\mathcal{L}_{\text{CE}}(\mathbf{z}^+,\hat{\mathbf{y}}) + \lambda \mathcal{L}_{\text{CE}}(1-\mathbf{z}^+,1-\hat{\mathbf{y}})$. The penalization parameter $\lambda\geq 0$ is difficult to tune in general.
This model realizes a blind homogeneous penalization of negative examples, independently of the image content,    
which encourages predicting only one positive label ($\sum_{l \in L}\hy_l\approx1$). 

We propose a trade-off between both approaches, with the weak negative loss
\begin{equation}\label{eq:lceoc}
     \mathcal{L}_{\lossname} = \mathcal{L}_{\text{CE}}(\mathbf{z}^+,\hat{\mathbf{y}}) + \mathcal{L}_{\text{CE}}(\tilde{\mathbf{z}}^-,1-\hat{\mathbf{y}}),
\end{equation}
where $\tilde{\mathbf{z}}^-$ is a weak estimation of the unknown ground truth negative examples. We highlight a main difference between our model and the one introduced in~\cite{cole2021multi}. In~\cite{cole2021multi},  all unobserved ground truth labels are learned, i.e. the unknown value of $\hat{y}_l$ is estimated  online for all $z^+_l=0$. On the other hand, we only aim at estimating partial weak negative examples $\tilde z_l^-=1$, corresponding to a subset of ground truth labels $\hat{y}_l=0$. In the next section, we detail how leveraging our image representation to obtain these negative examples.

\subsection{Negative example estimation with embedding self-similarities}\label{sec:neg}

In order to tackle the positive and unlabeled problem, we propose to estimate negative examples $\tilde{\mathbf{z}}=\{\tilde z_l^-\}_{l\in L}$ for each image. This estimation is done using the information contained in image representations $\Embi$, together with our initial postulate hypothesis:  at most one label can be observed in a patch. 

\subsubsection{Self similarity within image representations.}
We first show that two image representations, $\embi^{}_l$ and  $\embi^{}_k$ for labels $l$ and $k$ respectively, are similar if they come from similar image patches. To measure patch closeness,  we consider the cosine similarity metric $\operatorname{sim} (\vb*{u},\vb*{v}) = \frac{ \vb*{u} \; . \; \vb*{v} }{\lVert \vb*{u} \rVert_2 \times \lVert \vb*{v} \rVert_2}$ between two vectors $u$ and $v$. 
As defined in~\eqref{eq:attention-represention}, the image representation $\embi_l\in\Embi$ for label $l$ is designed to select patch embeddings $\embp_i\in \Embp$ that positively correlate with label embeddings $\embl_l$.
We also highlight that with our multi-class classifier,  label embeddings $\Embl=\{\embl_l,\, l\in L\}$ are assumed to cluster the embedding space. 
As a consequence, if the cosine similarity $\operatorname{sim}(\embi^{}_l, \embi^{}_k)$ between the image representations for two labels is large, it is most likely that these representations are based on similar subsets of patch embeddings $\embp_i$. Therefore, they come from similar image patches.

Next, as we assume that at most one label can be observed in a patch, if a set of patches is really representative of a label $l^*$, this set can not be relevant for characterizing other labels $k\neq l^*$. Thus, the model should not return a positive classification score for a label $k$  different from $l^*$.

Combining these observations, we conclude that when the cosine similarity $\operatorname{sim}(\embi^{}_l, \embi^{}_k)$ is large, 
the image representations $\embi^{}_l$ and $\embi^{}_k$ are based on the same set of patches and at least one of the  prediction for the labels $l$ and $k$ should be negative.

\subsubsection{Estimating negative labels.}
Based on this analysis, we propose to exploit self cosine similarities between image representations to estimate negative examples $\tilde{\mathbf{z}}^-$. 
We recall that at least one positive example, say $z_{l^*}^+=1$, is observed for any image. 
Hence, we rely on the value of the cosine similarity with observed labels, $\operatorname{sim}(\embi_{l^*}, \embi_k) \in [-1,1]$, to determine if unobserved labels $k$ can be considered as negative examples.
To that end, we first define the weights
\begin{equation}\label{neg_weights}
    \beta_{l,k}=\operatorname{\varphi}(
    \operatorname{sim}(\embi_l,\embi_k),\theta), 
\end{equation}
where $ \operatorname{\varphi}(x,\theta) = \mathbb{1}_{[x>\theta]}x$ is a thresholded Relu operator of parameter $\theta$. This parameter $\theta\in[-1,1]$ is the value at which the cosine similarity is small enough to consider the two embeddings $\embi_l$ and $\embi_k$ as different. Choosing $\theta\geq0$ guarantees that $\operatorname{\varphi}(\operatorname{sim}(\embi_l, \embi_k),\theta) \in [0,1]$.

\begin{wrapfigure}[14]{r}{0.385\textwidth}\vspace*{-0.6cm}
 \includegraphics[width=\linewidth]{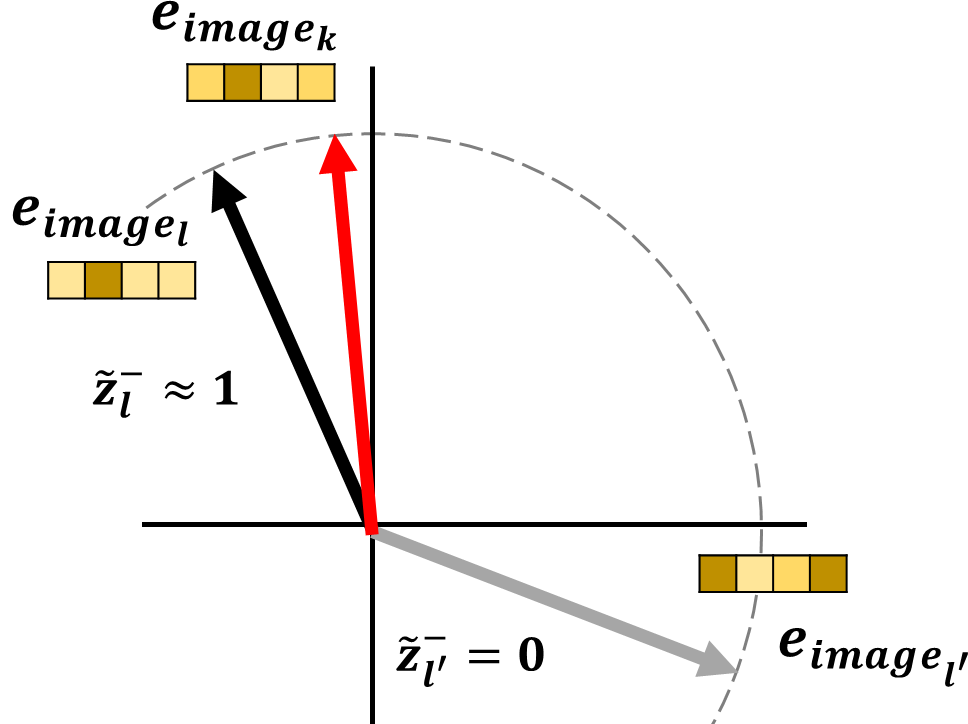}\vspace*{-0.1cm}
\caption{\label{fig:neg}Estimation of weak negative example scores $\tilde z_l^-$ and $\tilde z_{l'}^-$ for unobserved labels and $l$ and $l'$ from the single observed label $k$.}
\end{wrapfigure}Then, as illustrated in Fig. \ref{fig:neg}, we estimate for each image a negative example score $\tilde z_l^-$ for all unobserved labels $l \in L$  (i.e. when $z^+_l=0$):
\begin{equation}\label{eq:estneg}
    \widetilde{z}^-_l =  \max_{\substack{k\in L\\z_k^+=1}}%
    \beta_{l,k}.
\end{equation}
This states that the label $l$ can be considered as a (weak) negative example if its embedding is similar enough to the embedding of one of the observed labels $\zp_k=1$.
Negative scores $\widetilde{z}^-_l$ take continuous values in the range $[0,1]$ thus giving weak negative labels for $0<\widetilde{z}^-_l<1$.
The score is $0$ if the cosine similarity value is smaller than the threshold $\theta$ (see $\widetilde{z}^-_{l'}$ in Fig.~\ref{fig:neg}).\\

Reintroducing the image index $n$, and recalling the definition~\eqref{eq:lce} of the Cross Entropy, 
our weak negative loss function~\eqref{eq:lceoc}  can be rewritten as
\begin{align}
     \mathcal{L}_{\lossname} &= \sum_n\mathcal{L}_{\text{CE}}(\mathbf{z}^+_n,\hat{\mathbf{y}}_n) + \mathcal{L}_{\text{CE}}(\tilde{\mathbf{z}}^-_n,1-\hat{\mathbf{y}}_n) \nonumber\\
     &= - \sum_n\sum_{l \in L} 
     z^+_{n,l}\log(\hy_{n,l}) - \sum_n\sum_{l \in L} 
    \tilde z^-_{n,l}\log(1-\hy_{n,l}).
    \end{align}
This model thus provides negative  scores ${{z}}^-_{n,k}$ which values depend on the embedding similarity between unobserved labels ${z}^+_{n,k}=0$ and  observed ones ${z}^+_{n,l}=1$.


\section{Experiments}\label{sec:expes}
In this section, we first describe the experimental setting in section ~\ref{ssec:exp}.
In section~\ref{ssec:results}, we validate numerically our proposed architecture and our framework for negative estimation from multiple image representations self-similarities.
Comparisons and discussions are finally provided in section~\ref{ssec:discuss}.

\subsection{Settings}\label{ssec:exp}

\subsubsection{Datasets.}
Only few datasets propose multi-label annotated images. Datasets traditionally used for detection or segmentation can nevertheless be adapted to fit the multi-label learning problem. With such datasets, the ground truth bounding boxes or masks are discarded, and only the associated labels are conserved. All the instances of a label are considered as one label in the resulting annotation. Hence, in the ground truth annotation, $y_{n,l}=1$ means that at least one instance of label $l$ is observed in the image $x_n$. 
In our work, the single positive label is obtained with a uniform random selection between all ground truth labels.

We consider two datasets: COCO~\cite{lin2014microsoft} and Pascal VOC~\cite{everingham2010pascal}.
The 2012 version of VOC contains 5,717 images for model learning and 5,823 images for validation. Objects can be of 20 different classes, giving 20 different labels. Each color image can be at a resolution of up to $640\times640$. The 2014 version of COCO contains 82,081 images for model learning and 40,137 images for validation. The dataset is annotated with 80 different labels. The dataset images are also in colors and have a resolution of up to $500\times500$.


\subsubsection{Model architecture.}
The patch extraction was performed on $R=3$ downsized levels of resolution, with a factor of $d=2$ between each level. The patches are squared sub images of size $w=h=64$, extracted with a stride of $64$. For COCO and VOC, it results in a set of around $130$ patches per images. 

All embeddings $\embp_i$,  $\embl_l$ and $\embi_l$ are of size $F=256$. The $64\times64$ patches are processed with a CNN backbone. This patch embedder contains the first five
EfficientNet blocks with the same hyperparameters than~\cite{tan2019efficientnet}. The last two layers are composed of an average pooling layer and a fully connected layer of size $F=256$ in order to obtain the patch representations $\embp_i$.

The label embedding $\Embl$ of size $|L|\times F$ is learned by the model,  the number of labels being $|L|=80$ for COCO and $|L|=20$ for VOC. The attention pooling is performed with a regular cross attention and processed by a MLP of two layers with  $256$ neurons, resulting in $L$ image representations $\embi_l$ of size $F=256$. All weights are initialized with the unit variance scaling method. The GELU activation is used for the MLPs and the swish activation for all convolutions.

The experiments were conducted on one Tesla P40 GPU through an Azure virtual machine. 
Our model has been trained for $25$ epochs with batches of $16$ images. The starting learning rate is set to $lr=0.001$ and scheduled to decrease every 5 epochs, taking the different values $lr=[0.001,0.0005,0.00025,0.000125]$. We use the  optimizer LAMB \cite{you2019large} with a weight decay of $0.0001$. For the proposed negative estimation framework, we used $\theta=0$ to compute the weight $\beta$ in~\eqref{neg_weights}, resulting in a standard ReLU for the similarity normalization.

\subsubsection{Compared methods.}
We consider as reference the model trained using full supervision (i.e. all positives and negatives labels are known) with  $\mathcal{L}_{\text{BCE}}$~\eqref{eq:lbce}. We recall that all other models are trained using only one positive label per image.

We also compare our method with  two closely related methods~\cite{cole2021multi} and~\cite{verelst2022spatial} addressing the problem of multi-label  learning with single positive examples. 
All results are obtained using the mAP as the evaluation metric on the same validation sets of images and from models trained on the same datasets versions.
We highlight that Cole et al.~\cite{cole2021multi} present results obtained with a pre-trained (and fine-tuned) Resnet-50. 
Our architecture being trained from scratch, we therefore consider the setting of~\cite{verelst2022spatial}, obtained with conditions similar to ours. Hence, for comparisons with~\cite{cole2021multi}, we report the results from~\cite{verelst2022spatial}, where Resnet-50 architectures are trained from scratch with different losses during $100$ epochs.  

In practice, we provide comparisons with the models obtained after $100$ epochs with the losses $\mathcal{L}_{\text{AN}}$, $\mathcal{L}_{\text{EPR}}$, $\mathcal{L}_{\text{ROLE}}$ proposed in  \cite{cole2021multi} and $\mathcal{L}_{\text{EN+CL}}$, $\mathcal{L}_{\text{EN+SCL}}$ from \cite{verelst2022spatial}. In brief, $\mathcal{L}_{\text{AN}}$ is the "assume negative" loss considering all unobserved labels as negatives; $\mathcal{L}_{\text{EPR}}$ is the expected positive regularization applying a penalization to make the sum of positively predicted labels close to the average number of positive labels per image; and  $\mathcal{L}_{\text{ROLE}}$ is the regularized online label estimation which considers the same penalization and also estimates ground truth unobserved labels as parameters of the model for all images. Next,  $\mathcal{L}_{\text{EN+CL}}$ is the expected negative with consistency loss that uses augmented versions of an image; and $\mathcal{L}_{\text{EN+SCL}}$ is the expected negative with spatial consistency loss that predicts spatial classification scores on an augmented image feature map.

When training our architecture with $\mathcal{L}_{\text{EPR}}$, the average label per image is set to $k=2.92$ for COCO and $k=1.38$ for VOC. As recommended in \cite{cole2021multi}, for $\mathcal{L}_{\text{ROLE}}$, $lr$ is multiplied by $10$ in the branch estimating the value of all unobserved labels.

\vspace*{-6pt}
\subsection{Results and comparisons}\label{ssec:results}
\vspace*{-2pt}

We now present experiments to validate both the architecture and the framework for negative example estimation.
In Table \ref{tab:ownresults}, we provide mAP obtained with our architecture (top part) and the Resnet-50 one (bottom part).\\

For fair comparison of architectures, both models have been trained from scratch with the loss $\mathcal{L}_{\text{BCE}}$. 
This provides upper bounds for the architectures, as this experimental setting corresponds to the "ideal" case where all positive and negative are known.
Globally, our patch-based architecture seems adapted to multiple label learning. In full supervision, our architecture achieved $65.8$ on COCO, which is  better than the $64.8$ reported in~\cite{verelst2022spatial} for the Resnet-50. On VOC, the difference in favor of our patch model is significant ($61.6$ vs $53.4$).

To validate our negative example estimation framework, we present results obtained in the single positive case.
We recall that our strategy consists in using image representations self-similarities to estimate weak negative examples that are plugged into the  loss  $\mathcal{L}_{\lossname}$.
As illustrated in Table \ref{tab:ownresults}, for the same number of epochs, mAP results are close to the ones obtained with full supervision  ($63.2$ vs $65.8$ for COCO and $60.4$ vs $61.6$ for VOC).\\

We also trained our patch-based architecture with the competing $\mathcal{L}_{\text{AN}}$, $\mathcal{L}_{\text{ROLE}}$ and  $\mathcal{L}_{\text{EPR}}$ losses. 
Our architecture trained with our negative estimation proposition gives the best results both on COCO and VOC. 
With $\mathcal{L}_{\text{AN}}$, the sum of predicted labels is always close to $1$, which better fits the VOC dataset (which true mean number of labels per image is $k=1.38$), than the COCO one ($k=2.92$).
It should be noticed that the $\mathcal{L}_{\text{ROLE}}$ underperforms with the considered learning setting. We suggest this is due to the complexity of the model, which intends to estimate all unobserved labels  with a dedicated branch in the loss function.\\
%

\setlength{\tabcolsep}{8pt}
\begin{table}[!t]
\centering%
\begin{tabular}{clcc}
\toprule
Model & Loss/method & COCO-14 & VOC-12 \\
\midrule 
\multirow{4}{*}{\begin{tabular}{c}Patch based \\architecture \\(ours)\end{tabular}} & $\mathcal{L}_{\text{BCE}}$ (fully-annotated) & 65.8 & 61.6 \\
\cmidrule(l){2-4}
 & $\mathcal{L}_{\text{AN}}$~\cite{cole2021multi} & 62.6 & 60.0 \\
 & $\mathcal{L}_{\text{EPR}}$~\cite{cole2021multi} & 61.4 & 58.8 \\
 & $\mathcal{L}_{\text{ROLE}}$~\cite{cole2021multi} & 33.3 & 49.7 \\
 & $\mathcal{L}_{\lossname}$ (ours)& \textbf{63.2} & \textbf{60.4}\\
\midrule 
\multirow{4}{*}{\begin{tabular}{c}Resnet-50 \\(reported in \cite{verelst2022spatial})\end{tabular}} & $\mathcal{L}_{\text{BCE}}$ (fully-annotated) & 64.8 & 53.4 \\ 
\cmidrule(l){2-4}
& $\mathcal{L}_{\text{AN}}$~\cite{cole2021multi} & 50.2 & 45.7 \\
& $\mathcal{L}_{\text{ROLE}}$~\cite{cole2021multi} & 51.9 & 45.0\\
& $\mathcal{L}_{\text{EN+CL}}$~\cite{verelst2022spatial} & 54.3 & 47.0 \\
& $\mathcal{L}_{\text{EN+SCL}}$~\cite{verelst2022spatial} & 54.0 & 50.4 \\
\bottomrule
\end{tabular}
\caption{mAP results for our patch approach trained for $25$ epochs and different losses (top) and comparisons (bottom) with the results reported by \cite{verelst2022spatial} after training a Resnet-50 architecture for $100$ epochs. Best results in bold. }%
\label{tab:ownresults}
\vspace*{-5pt}
\end{table}
For completeness, we reproduce in Table \ref{tab:ownresults} (bottom part) the mAP results reported in~\cite{verelst2022spatial}, when training a Resnet-50 architecture from scratch with the competing losses. 
Contrary to our architecture, the decrease of mAP performance with single positive examples is significant with respect to the full supervision. 

All these results indicate that the proposed framework is adapted to the complex multi-label PU problem including only single positive examples.\vspace{-0.2cm}

\subsection{Discussion}\label{ssec:discuss}

\subsubsection{Computational burden.} The computational burden for training our patch-based architecture is significantly reduced with respect to the Resnet-50 model of ~\cite{cole2021multi} and~ \cite{verelst2022spatial}.
First, the model size is reduced by a factor $100$. Our model has approximately $250K$ parameters to learn, whereas the Resnet-50 architecture is composed of $23M$ parameters.
With our light patch-based architecture,  better results are also obtained with only $25$ epochs, instead of $100$ epochs for the Resnet-50 architectures. 
This suggests that our architecture generalizes better.


\subsubsection{Hyper-parameter.} Our weak negative loss does not include any hyper-parameter. Excepting the network architecture, the only extra hyper-parameter of our full model is the threshold of the cosine similarity in  \eqref{neg_weights} that we simply fix to $\theta=0$. On the other hand, 
the losses $\mathcal{L}_{\text{EPR}}$ and  $\mathcal{L}_{\text{ROLE}}$  penalize the sum of positive predictions with respect to the average number of label per image. 
This is a strong assumption, as the variance of positive labels on all image of the dataset can be large. Moreover, the prior knowledge of the mean number of labels  is often not available in real use cases. 
The loss $\mathcal{L}_{\text{ROLE}}$ also relies on an online estimation of all unobserved ground truth annotations from current predictions. This model is thus greatly influenced by the initialization and the first few epochs, while increasing the memory requirements.

\setlength{\tabcolsep}{1pt}
\begin{figure}[!t]
\centering%
\begin{tabular}{ccccc}
    \includegraphics[width=0.19\linewidth]{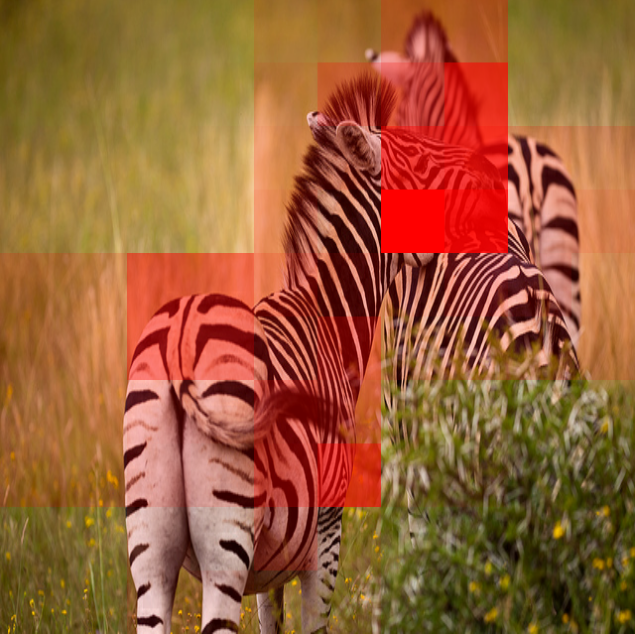} &
    \includegraphics[width=0.19\linewidth]{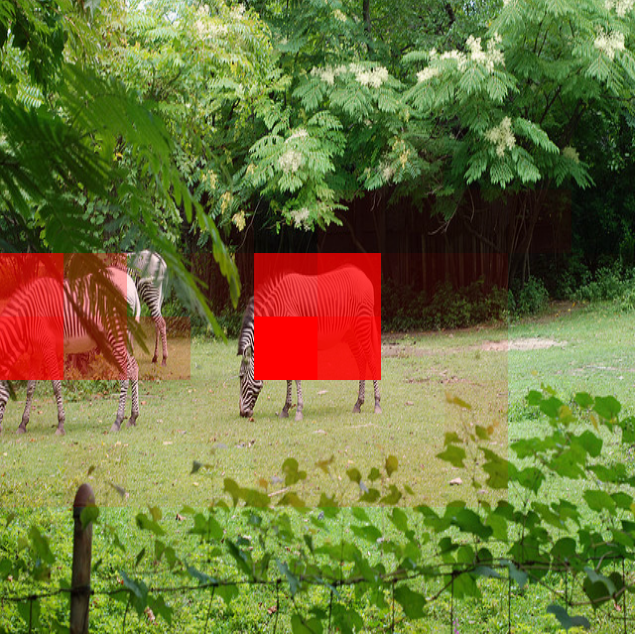} &
    \includegraphics[width=0.19\linewidth]{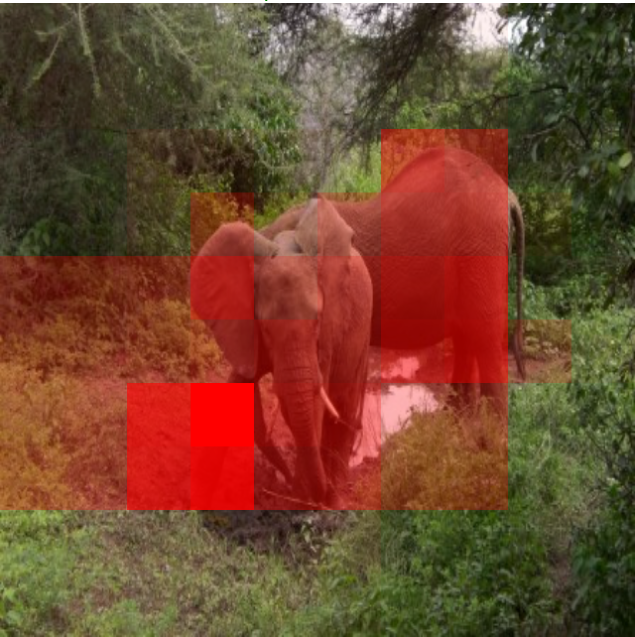} &
    \includegraphics[width=0.19\linewidth]{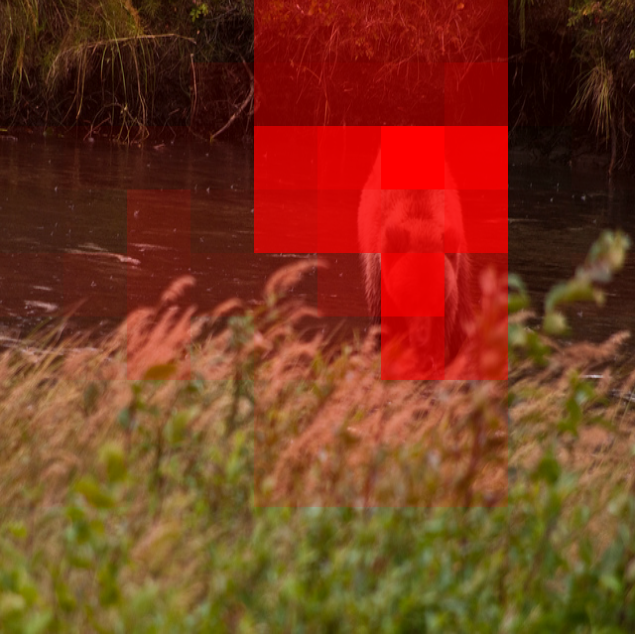} &
    \includegraphics[width=0.19\linewidth]{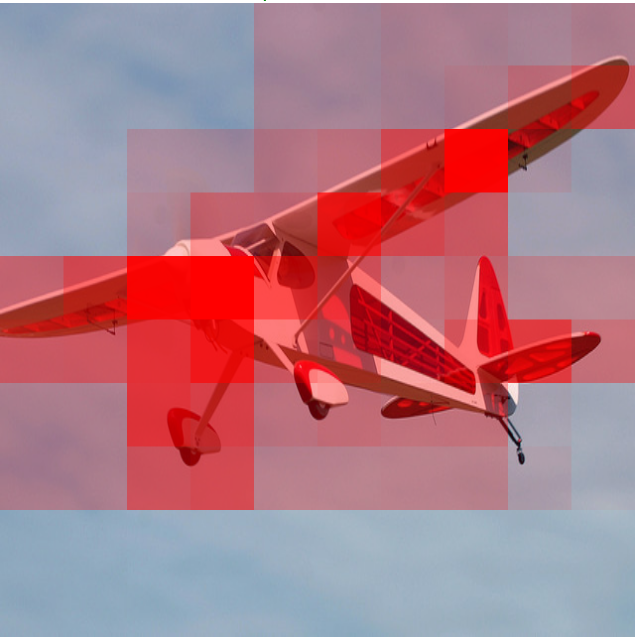}\\
     zebra: 94.8 & zebra: 98.9 & elephant: 58.6 & bear: 57.1 & airplane: 77.5
\end{tabular}
\caption{Examples of patch attention scores for true positives. Patches are filled with their attention score values, that vary from $0$ (transparent) to $1$ (red).  The second line present the prediction score for the given labels.}
\label{fig:attention-score-img}
\vspace*{-5pt}
\end{figure}
\subsubsection{Label localization.} Our architecture has the potential to locate patch examples corresponding to a detected label (Fig.~\ref{fig:attention-score-img}). Indeed, the attention scores computed in the attention patch pooling  (block 4 of Fig.~\ref{fig:archi}) allows determining the level of patch participation to the classification decision. The patch-based model thus offers a natural framework to interpret the obtained results, without relying on advanced gradient backpropagation mechanisms~\cite{selvaraju2017grad}.

\subsubsection{Pre-training and fine-tuning.} It is important to underline the current limitation of our approach with respect to Resnet-50 architectures. The models \cite{cole2021multi} and \cite{verelst2022spatial} provide significant better mAP results  ($72$ for COCO and even $88$ for VOC), when considering a Resnet-50 pre-trained on Imagenet, with potential fine-tuning refinements. We postulate that our  results could also be improved by pretraining either our full patch-based  architecture on Imagenet, or the patch embedder on bounding boxes of a detection dataset. The performance could also be increased by conducting an extensive hyper-parameter search (dimension of the representation space $F$, number of attention layers, ...) and training tuning. 

\section{Conclusion}

In this work, we proposed a light patch-based architecture for multi-label classification problems. Leveraging on patch embedding self-similarities, we provide a strategy for estimating negative examples when facing the challenging problem of positive and unlabeled learning. The patch-based attention strategy also gives a natural framework to localize detected labels within images.

Numerical experiments demonstrate the interest of the approach when no dedicated pre-trained network is available. Our model is able to generalize fast from few labels, as it provides relevant results when trained from scratch during a few epochs. In the related literature, the best  performances are currently obtained with  pre-trained Resnet-50 architectures having a number of parameters $100$ times greater than ours.

In order to improve our model and reach state-of-the-art results, the main directions we draw are the online estimation of positive and negative example at batch level, the pre-training of the patch embedder and an improved model to cluster the patch embedding space with respect to the labels.

\bibliographystyle{abbrv}



\end{document}